\documentclass{article}

\usepackage[english]{babel}

\usepackage[letterpaper,top=2cm,bottom=2cm,left=3cm,right=3cm,marginparwidth=1.75cm]{geometry}

\usepackage[utf8]{inputenc} 
\usepackage[T1]{fontenc}    
\usepackage{hyperref}       
\usepackage{url}            
\usepackage{booktabs}       
\usepackage{amsfonts}       
\usepackage{nicefrac}       
\usepackage{microtype}      
\usepackage{xcolor}         
\usepackage{wrapfig}
\usepackage{multicol}
\usepackage{amsmath}
\usepackage{amssymb}
\usepackage{mathtools}
\usepackage{amsthm}
\usepackage{fdsymbol}
\usepackage{tabularx}
\usepackage[ruled,linesnumbered]{algorithm2e}

\usepackage{algorithmic}

\theoremstyle{plain}
\newtheorem{theorem}{Theorem}[section]

\newtheorem{corollary}[theorem]{Corollary}
\theoremstyle{definition}
\newtheorem{definition}[theorem]{Definition}

\theoremstyle{remark}

\usepackage{natbib}
\usepackage{authblk}
\usepackage{tikz}
\usetikzlibrary{bayesnet}
\usetikzlibrary{arrows.meta}
\usepackage{subfigure}

\tikzstyle{causallatent} = [circle,fill=white,draw=black,inner sep=1pt,
minimum size=18pt, font=\fontsize{10}{10}\selectfont, node distance=0.1]
\tikzstyle{symbol} = [circle,fill=white,draw=white,inner sep=1pt,
minimum size=18pt, font=\fontsize{10}{10}\selectfont, node distance=0]
\tikzstyle{causalobs} = [causallatent,fill=gray!25, node distance=0.8]

\tikzstyle{causaleps} = [rectangle,fill=white,draw=black,inner sep=1pt,
minimum size=18pt, font=\fontsize{8}{8}\selectfont, node distance=1]

\tikzstyle{causalobsset} = [ellipse,fill=white,draw=black,inner sep=1pt,
minimum size=18pt, font=\fontsize{8}{8}\selectfont, node distance=1, fill=gray!25]

\tikzstyle{node} = [circle,fill=gray!25,draw=black,inner sep=1pt,
minimum size=14pt, font=\fontsize{10}{10}\selectfont, node distance=0.6]

\tikzstyle{ellipsenode} = [ellipse,fill=gray!25,draw=black,inner sep=1pt,
minimum size=18pt, minimum width=60pt, font=\fontsize{10}{10}\selectfont, node distance=0.6]

\title{Identification of Causal Direction under an Arbitrary Number of Latent Confounders}

\author[1]{Wei Chen}
\author[1]{Linjun Peng}
\author[1]{Zhiyi Huang}
\author[2]{Haoyue Dai}
\author[1,3]{Zhifeng Hao}
\author[1,4]{Ruichu Cai}
\author[2,5]{Kun Zhang}
\affil[1]{School of Computer Science, Guangdong University of Technology, Guangzhou, China}
\affil[2]{Department of Philosophy, Carnegie Mellon University, Pittsburgh, PA, United States}
\affil[3]{College of Engineering, Shantou University, Shantou, China}
\affil[4]{Peng Cheng Laboratory, Shenzhen, China}
\affil[5]{Mohamed bin Zayed University of Artificial Intelligence, Abu Dhabi, United Arab Emirates}

\date{}

\begin{document}

\maketitle

\begin{abstract}
Recovering causal structure in the presence of latent variables is an important but challenging task. While many methods have been proposed to handle it, most of them require strict and/or untestable assumptions on the causal structure. In real-world scenarios, observed variables may be affected by multiple latent variables simultaneously, which, generally speaking, cannot be handled by these methods. In this paper, we consider the linear, non-Gaussian case, and make use of the joint higher-order cumulant matrix of the observed variables constructed in a specific way. We show that, surprisingly, causal asymmetry between two observed variables can be directly seen from the rank deficiency properties of such higher-order cumulant matrices, even in the presence of an arbitrary number of latent confounders. Identifiability results are established, and the corresponding identification methods do not even involve iterative procedures. Experimental results demonstrate the effectiveness and asymptotic correctness of our proposed method.

\end{abstract}

\section{Introduction}

Causal discovery aims to reveal the underlying causal structure among variables, which has become increasingly important in applications across various scientific disciplines. In real-world scenarios, not all relevant causal variables can be observed or measured. The presence of latent variables that simultaneously affect multiple observed variables, poses significant challenges to the identifiability of causal structures~\citep{spirtes2000causation,hyvarinen2013pairwise,chen2021causal,adams2021identification}. 
To address this problem, researchers have developed several methods, primarily categorized into constraint-based, score-based, and functional-based ones. While constraint-based~\citep{spirtes2000causation,kaltenpoth2023causal} and score-based methods~\citep{jabbari2017discovery,ngscore2024} have made significant contributions, they often suffer from the Markov Equivalence Class problem and cannot uniquely identify causal directions. 

With specific assumptions on the data generating process, Latent variables Linear Non-Gaussian Model (LvLiNGAM)~\cite{hoyer2008estimation} is one of the typical functional-based methods for causal discovery in the presence of hidden confounders, with several variants. 
One prominent approach is based on Overcomplete Independent Component Analysis (OICA)~\cite{eriksson2004identifiability,lewicki2000learning}, which estimates the mixing matrix first and then transforms it into the causal strength matrix. But in practice, OICA is computationally challenging and is often prone to local optima~\cite{naik2011overview}, thus limiting its practical applicability. The problem of identifying causal relationships among observed variables under an arbitrary number of latent confounders remains largely unsolved.
 
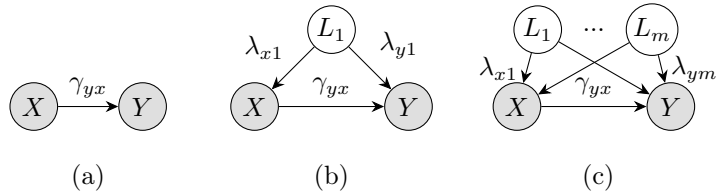
\begin{figure*}[t]
    \centering
    \begin{minipage}[b]{0.2\textwidth}
        \centering
        \begin{tikzpicture}
            \node[causalobs] (X) {$X$};
            \node[causalobs, right = of X] (Y) {$Y$};
            
            \draw[-{Stealth[width=1.5mm]}] (X) -- (Y) node[midway, above] {$\gamma_{yx}$};
        \end{tikzpicture}
        \begin{center}
            (a)
        \end{center}
    \end{minipage}
    \begin{minipage}[b]{0.2\textwidth}
        \centering
        \begin{tikzpicture}
            \node[causallatent] (L1) {$L_1$};
            \node[causalobs, below left = of L1] (X) {$X$};
            \node[causalobs, below right = of L1] (Y) {$Y$};
            
            \draw[-{Stealth[width=1.5mm]}] (L1) -- (Y) node[midway, above right=0mm and 0mm] {$\lambda_{y1}$};
            \draw[-{Stealth[width=1.5mm]}] (L1) -- (X) node[midway, above left=0mm and 0mm] {$\lambda_{x1}$};
            \draw[-{Stealth[width=1.5mm]}] (X) -- (Y) node[midway, above] {$\gamma_{yx}$};
    
        \end{tikzpicture}
        \begin{center}
            (b)
        \end{center}
    \end{minipage}
    \begin{minipage}[b]{0.24\textwidth}
        \centering
        \begin{tikzpicture}
            \node[symbol] (L) {$...$};
            \node[causallatent, left = of L] (L1) {$L_1$};
            \node[causallatent, right = of L] (Lm) {$L_m$};
            \node[causalobs, below left = of L] (X) {$X$};
            \node[causalobs, below right = of L] (Y) {$Y$};
            
            \draw[-{Stealth[width=1.5mm]}] (L1) -- (Y) ;
            \draw[-{Stealth[width=1.5mm]}] (L1) -- (X) node[midway, left=0mm] {$\lambda_{x1}$};
            \draw[-{Stealth[width=1.5mm]}] (Lm) -- (Y) node[midway, right=0mm] {$\lambda_{ym}$};
            \draw[-{Stealth[width=1.5mm]}] (Lm) -- (X) ;
            \draw[-{Stealth[width=1.5mm]}] (X) -- (Y) node[midway, above] {$\gamma_{yx}$};
        \end{tikzpicture}
        \begin{center}
            (c)
        \end{center}
    \end{minipage}
\caption{Three causal structures over two observed variables $X$ and $Y$ with causal relationship $X \to Y$, each influenced by a different number of latent variables. (a) No latent variables. (b) One latent variable $L_1$. (c) Multiple latent variables $L_1, \dots, L_m$.}
\label{fig:data generating processes}
\end{figure*}

In this context, higher-order cumulants emerge as a powerful tool for capturing essential statistical properties, particularly for non-Gaussian distributions. Recent years have seen growing interest in leveraging cumulants for causal discovery. Robeva and Seby~\cite{robeva2021multi} explore interactions among multiple sets of random variables in a linear structural equation model with non-Gaussian errors. Transformed Independent Noise (TIN) condition \cite{dai2022independence} is proposed to identify the ordered group decomposition of the causal model. A closed-form solution is proposed for discovering the one-latent component structure \cite{cai2023causal}. The aforementioned methods require information from at least three observed variables. For situations involving only two variables, some cumulant-based criteria \cite{chen2024identification} are proposed to infer the causal direction between the two observed variables that are affected by a single latent variable, as shown in Fig. \ref{fig:data generating processes} (b). In a special case, if multiple latent variables have perfectly collinear effects, they can be mathematically treated as a single latent variable. However, in most real-world scenarios, multiple latent variables influence observed variables in ways that are not collinear (i.e. Fig. \ref{fig:data generating processes} (b) and Fig. \ref{fig:data generating processes} (c) are not equivalent). Thus, the existing method \cite{chen2024identification} cannot be extended to handle the problem of inferring causal relationships among observed variables in the presence of an arbitrary number of latent confounders, which is given in Fig.~\ref{fig:data generating processes} (c).

To tackle the above problem, we find that, surprisingly, utilizing more terms of higher-order joint cumulants of observed variables enables us to extract richer information from non-Gaussian data. Those different terms of higher-order joint cumulants can be regarded as the elements of a matrix, which is denoted as \textit{cumulant matrix}. Interestingly, by comparing the rank of the cumulant matrix, we can directly uncover the inherent asymmetry in causal relationships between variables, without even resorting to iterative procedures. Thus, we propose a practical causal discovery algorithm based on the higher-order cumulant matrix to handle the case with an arbitrary number of latent variables. We demonstrate that the number of latent variables and the causal relationship between two observed variables can be identified based on the rank constraint of the proposed joint cumulant matrix, through using the determinant of the matrix. Our contributions are as follows: 1) we established a more general identification condition under multiple latent confounders, and provided theoretical guarantees for identifying causal direction between two observed variables that are affected by an arbitrary number of latent confounders; 2) we provided a novel method for causal discovery under an arbitrary number of latent confounders; 3) the experimental results show the effectiveness of our proposed method.

\section{Preliminaries}

\subsection{Latent-Variable Linear Non-Gaussian Acyclic Model}

In this paper, we consider data over observed variables $X$ and $Y$, which may be influenced by latent variables $\mathbf{L} = \{L_1, ..., L_m\}$. Specifically, assume $X$ is the cause of $Y$, then the data is generated by a linear causal model with non-Gaussian noise, which can be formalized as
\begin{equation}\label{eq:LvLiNGAM}
    \begin{aligned}
         L_i &= E_{li},\\
         X &=  \sum_{i=1}^{m}\lambda_{xi} L_i + E_{x},\\
         Y &= \gamma_{yx} X + \sum_{i=1}^{m}\lambda_{yi} L_i + E_{y},
    \end{aligned}
\end{equation}
where $\gamma_{yx}$ is the causal coefficient from $X$ to $Y$. $\lambda_{xi}, \lambda_{yi}$ represents the causal coefficient from the latent variable $L_i$ to $X, Y$, respectively. $E_{li}$, $E_{x}$ and $E_{y}$ are the independent noise terms for $L_i$, $X$ and $Y$, respectively.
This model is also termed Latent-Variable Linear Non-Gaussian Acyclic Model (LvLiNGAM)~\cite{hoyer2008estimation}.

We can directly transform the above LvLiNGAM model into a form of path coefficients and independent noise terms, which can be formalized as:
\begin{equation}
    \begin{aligned}
        X & = \sum_{i=1}^{m} \alpha_i E_{li} + E_x,\\
        Y & = \gamma_{yx} E_x + \sum_{i=1}^{m} \beta_i E_{li} + E_y,
    \end{aligned} 
\label{eq:path coefficients LvLiNGAM X,Y}
\end{equation}
where $\alpha_i$ and $\beta_i$ represent the total effects of the latent variable $L_i$ on $X$ and $Y$, respectively. Considering the irreducible, we assume that if at least one latent variable affects $X$ and $Y$, then $\alpha_i, \beta_i \neq 0$ in this paper. Please refer to Appendix for more elaboration. 

\subsection{Cumulants and its Role in Capturing Distributional Information}

We briefly outline the cumulants and describe its role in characterizing distributional information. The definition of cumulants~\cite{brillinger2001time} is as follows: 
\begin{definition}
    [\textbf{Cumulants}~\cite{brillinger2001time}]
Let $\mathbf{Z}=(Z_1, Z_2, \dots, Z_n)$ be a random vector of length $n$. The $k$-th order cumulant tensor of $\mathbf{Z}$ is defined as a $n \times \cdots  \times n$ ($k$ times) table, $\mathcal{C}^{(k)}$, whose entry at position ($i_1, \cdots, i_k$) is
\begin{align*}
    \mathcal{C}^{(k)}_{i_1 \!, \cdots \!,  i_k} = \text{Cum}(Z_{i_1}, \dots, Z_{i_k}) 
    =  \sum_{(d_1\!, \dots \!, d_h)} (-1)^{h \!- \!1}(h \!- \!1)! \mathbb{E}\!\left[ \! \prod_{j\in d_i}Z_j  \! \right] \cdots \mathbb{E} \!\left[ \! \prod_{j \in d_h} \! Z_j \! \right]\!,
\end{align*}
where the sum is taken over all partitions $(d_1, \dots, d_h)$ of the set $\{i_1, \dots, i_k\}$. 
\end{definition}

For convenience, we use $\mathcal{C}_{a}(Z_i)$ to denote the cumulant $\text{Cum}(\underbrace{Z_i, \dots}_{a \,\text{times}})$, and use $\mathcal{C}_{a, b}(Z_i, Z_j)$ to denote the joint cumulant $\text{Cum}(\underbrace{Z_i, \dots}_{a \,\text{times}}, \underbrace{Z_j, \dots}_{b \,\text{times}})$. For example, $\mathcal{C}_{3}(Z_i)$ represents $\text{Cum}(Z_i, Z_i, Z_i)$, $\mathcal{C}_{1, 2}(Z_i, Z_j)$ represents $\text{Cum}(Z_i, Z_j, Z_j)$. 

Note that the first-order cumulant of $Z_i$ is the mean of $Z_i$, and the second-order cumulant of $Z_i$ is the variance of $Z_i$. While the first and second-order information is sufficient to describe a Gaussian distribution, the information in a non-Gaussian distribution can be obtained by higher-order cumulants. The third-order cumulant (skewness) and fourth-order cumulant (kurtosis) are important and widely used measures of non-Gaussianity~\cite{hyvarinen2001independent}. For some non-Gaussian distributions, while certain order cumulants may not exist or equal zero, there will always be at least one higher-order cumulant (above second-order) do exists. These existing higher-order cumulants can be used to characterize the distribution's inherent properties. 

\section{Intuition: Cumulants Contain Information of Causal Asymmetry} 

Let us start by providing the intuition about how to see causal asymmetry from a properly constructed cumulant. We start with the unconfounded case and then extend it to the case with latent confounders. Consider the causal graph over two observed variables $X$ and $Y$ given in Fig.~\ref{fig:data generating processes} (a). 
Suppose that data over $X$ and $Y$ is generated by Eq. (\ref{eq:LvLiNGAM}). Then, the joint cumulants $\mathcal{C}_{a,b}(X, Y)$ can be expressed as
\begin{align*}
    \mathcal{C}_{a,b}(X, Y) = \alpha_{x}^{a}\beta_{x}^{b}\mathcal{C}_{a+b}(E_x) + \alpha_{y}^{a}\beta_{y}^{b}\mathcal{C}_{a+b}(E_y),
\end{align*}
where in Fig. \ref{fig:data generating processes} (a), $\alpha_{x}$, $\beta_{y} = 1$, and $\alpha_{y} = 0, \beta_{x}=\gamma$; $a, b \in \mathbb{Z}_{\geq 0}$, $a+b=3$. 
We find that the joint cumulants of $X$ and $Y$ can be regarded as the sum of the multiplication of powers of the path coefficient and the cumulant of the noise terms. 
Now we are ready to see the asymmetry between $X$ and $Y$. 
Specifically, we construct the joint cumulant matrix of $X$ and $Y$, denoted as $CM^{(2)}_{(X,Y)}$ and $CM^{(2)}_{(Y,X)}$, as follows:
\begin{equation}
\begin{aligned}
    CM^{(2)}_{(X,Y)} &= 
    \begin{bmatrix}
    \mathcal{C}_{3,0} (X,Y) & \mathcal{C}_{2,1} (X,Y)\\
    \mathcal{C}_{2,1} (X,Y) & \mathcal{C}_{1,2} (X,Y)
    \end{bmatrix}= 
    \begin{bmatrix}
    \mathcal{C}_3(E_x) & \gamma\mathcal{C}_3(E_x)\\
    \gamma\mathcal{C}_3(E_x) & \gamma^2\mathcal{C}_3(E_x)
    \end{bmatrix},\\
    CM^{(2)}_{(Y,X)} &= 
    \begin{bmatrix}
    \mathcal{C}_{3,0} (Y,X) & \mathcal{C}_{2,1} (Y,X)\\
    \mathcal{C}_{2,1} (Y,X) & \mathcal{C}_{1,2} (Y,X)
    \end{bmatrix}= 
    \begin{bmatrix}
    \gamma^3\mathcal{C}_3(E_x) + \mathcal{C}_3(E_y) & \gamma^2\mathcal{C}_3(E_x)\\
    \gamma^2\mathcal{C}_3(E_x) & \gamma\mathcal{C}_3(E_x)
    \end{bmatrix}.
\end{aligned}
\label{eq:intuition}
\end{equation}
From Eq.~\eqref{eq:intuition}, we observe that $CM^{(2)}_{(X,Y)}$ exhibits proportionality in its rows and columns and hence rank deficiency, whereas $CM^{(2)}_{(Y,X)}$ has no such proportionality due to the additional term $\mathcal{C}_3(E_y)$. 

Now let us consider more interesting cases, with latent confounders. Take the causal structure given in Fig.~\ref{fig:data generating processes} (b) and (c) as examples, we can observe the rank discrepancy of two cumulant matrices. For two observed variables affected by latent variables, the independent noise terms within their joint cumulants increase accordingly. This phenomenon induces more intricate proportional relationships within the cumulant matrix. However, the intrinsic relationships remain governed by the proportional coefficients of path connections. These proportional characteristics can be intuitively demonstrated through a specific matrix factorization formulation. For the data generated according to Fig.~\ref{fig:data generating processes} (b),
\begin{align*}
    CM^{(3)}_{(X,Y)} &= 
    \begin{bmatrix}
    \mathcal{C}_{5,0} (X,Y) & \mathcal{C}_{4,1} (X,Y) & \mathcal{C}_{3,2} (X,Y) \\
    \mathcal{C}_{4,1} (X,Y) & \mathcal{C}_{3,2} (X,Y) & \mathcal{C}_{2,3} (X,Y) \\
    \mathcal{C}_{3,2} (X,Y) & \mathcal{C}_{2,3} (X,Y) & \mathcal{C}_{1,4} (X,Y) \\
    \end{bmatrix},\\
    CM^{(3)}_{(Y,X)} &= 
    \begin{bmatrix}
    \mathcal{C}_{5,0} (Y,X) & \mathcal{C}_{4,1} (Y,X) & \mathcal{C}_{3,2} (Y,X) \\
    \mathcal{C}_{4,1} (Y,X) & \mathcal{C}_{3,2} (Y,X) & \mathcal{C}_{2,3} (Y,X) \\
    \mathcal{C}_{3,2} (Y,X) & \mathcal{C}_{2,3} (Y,X) & \mathcal{C}_{1,4} (Y,X) \\
    \end{bmatrix}.
\end{align*}
we can express the matrix factorization formulation of cumulant matrix $CM^{(3)}_{(X,Y)}$, $CM^{(3)}_{(Y,X)}$ as:
\begin{align*}
    CM^{(3)}_{(X,Y)} = CM^{(path)}_{(X,Y)} CM^{(source)}_{(X,Y)} CM^{(path)\text{T}}_{(X,Y)},\\
    CM^{(3)}_{(Y,X)} = CM^{(path)}_{(Y,X)} CM^{(source)}_{(Y,X)} CM^{(path)\text{T}}_{(Y,X)},
\end{align*}
where $CM^{(path)}_{(X,Y)}$ and $CM^{(source)}_{(Y,X)}$ are as follows:
\begin{align*}
    CM^{(path)}_{(X,Y)} &= CM^{(path)}_{(Y,X)} = 
    \begin{bmatrix}
    \alpha_{1}^{2} \beta_{1}^{0} & 1 & 0\\
    \alpha_{1}^{1} \beta_{1}^{1} & \gamma^{1} & 0\\
    \alpha_{1}^{0} \beta_{1}^{2} & \gamma^{2} & 1\\
    \end{bmatrix},\\
    CM^{(source)}_{(X,Y)} &= 
    \begin{bmatrix}
    \alpha_{1} \mathcal{C}_{5}(E_{l1}) & 0 & 0\\
    0 & 1 \times \mathcal{C}_{5}(E_{x}) & 0\\
    0 & 0 & 0 \times \mathcal{C}_{5}(E_{y})\\
    \end{bmatrix},\\
    CM^{(source)}_{(Y,X)} &= 
    \begin{bmatrix}
    \beta_{1} \mathcal{C}_{5}(E_{l1}) & 0 & 0\\
    0 & \gamma \times \mathcal{C}_{5}(E_{x}) & 0\\
    0 & 0 & 1 \times \mathcal{C}_{5}(E_{y})\\
    \end{bmatrix}.
\end{align*}
Intuitively, the rank of the cumulant matrix can represent the number of independent noises in $X$ and $Y$, respectively, and this results in a rank discrepancy between two cumulant matrices: $2= rank(CM^{(3)}_{(X,Y)})<rank(CM^{(3)}_{(Y,X)}) = 3$. Inspired by this, we leverage the rank of the cumulant matrix defined above (depending on the number of latent variables) to identify the causal direction between two observed variables, despite the influence of any number of latent variables.

\section{Proposed Method}

In this section, we first define the constructed cumulant matrix and then explore its benefit for identifying causal relationships among observed variables without latent variables. By increasing the order of the cumulant matrices, we propose identifiability theories to account for latent variables. Finally, we introduce a practical approach to determine causal directions between observed variables, accommodating an arbitrary number of latent variables.

\subsection{Cumulant Matrix}

\begin{definition}\label{df:Construct the cumulant matrix.}
\textbf{Cumulant matrix.}
Assume that the data over $ X $ and $ Y $ are generated by Eq.~\eqref{eq:path coefficients LvLiNGAM X,Y}, the $k\times k$ cumulant matrix $CM^{(k)}_{(X,Y)}$ is defined as follows:
\begin{align*}
    CM^{(k)}_{(X,Y)} =  \begin{bmatrix}
        \mathcal{C}_{2k-1,0}(X,Y) & \mathcal{C}_{2k-2,1}(X,Y) & \cdots & \mathcal{C}_{k,k-1}(X,Y) \\
        \mathcal{C}_{2k-2,1}(X,Y) & \mathcal{C}_{2k-3,2}(X,Y) & \cdots & \mathcal{C}_{k-1,k}(X,Y) \\
        \vdots & \vdots & \ddots & \vdots \\
        \mathcal{C}_{k,k-1}(X,Y) & \mathcal{C}_{k-1,k}(X,Y) & \cdots & \mathcal{C}_{1,2k-2}(X,Y)
    \end{bmatrix}.
\end{align*}
Each element of the cumulant matrix $CM^{(k)}_{(X,Y)}$ in the $ i $-th row and $ j $-th column, denoted as joint cumulant $\mathcal{C}_{a_i,b_j}(X,Y)$. The indices $ a_i $ and $ b_j $ are defined based on the position of the element within the matrix as $a_i = 2k + 1 - i - j$ and $b_j = i + j - 2$.

Intuitively, the orders of joint cumulants between adjacent rows and columns in the cumulant matrix differ by one. This enables us to more effectively leverage the linear separability of cumulants under independent noise to characterize causal relationships among variables. For clarify, the rank of $CM^{(k)}_{(X,Y)}$ is represented by $rank(CM^{(k)}_{(X,Y)})$ and the determinant of the matrix $CM^{(k)}_{(X,Y)}$ is denoted as $|CM^{(k)}_{(X,Y)}|$, the absolute value of $|CM^{(k)}_{(X,Y)}|$ is defined as $||CM^{(k)}_{(X,Y)}||$.
\end{definition}

\subsection{Causal Discovery without Latent Variables}

Based on the proposed cumulant matrices, we provide the following theorem to identify the causal relationship between $X$ and $Y$ that are not affected by latent variables.
\begin{theorem}\label{th:causal discovery of observed variable}
Assume that $ X $ and $ Y $ are generated by Eq.~\eqref{eq:path coefficients LvLiNGAM X,Y} and there are no shared latent variables influencing them. Define $ CM^{(k)}_{(X,Y)} $ and $ CM^{(k)}_{(Y,X)} $ as cumulant matrices of $ X $ and $ Y $ with $k = 2$. Then, 1) $X$ is a cause of $Y$ if and only if $rank(CM^{(k)}_{(X,Y)})<rank(CM^{(k)}_{(Y,X)})$; 2) $X$ and $Y$ are mutually independent if and only if $rank(CM^{(k)}_{(X,Y)})=rank(CM^{(k)}_{(Y,X)})=1$.
\end{theorem}

Theorem~\ref{th:causal discovery of observed variable} guarantees that the rank deficiency of the cumulant matrix can identify the causal structure between two observed variables without the influence of latent variables. 

\subsection{Causal Discovery with Latent Variables}

Given the data over $X$ and $Y$ are generated by Eq.~\eqref{eq:path coefficients LvLiNGAM X,Y}. If there is one or more than one latent variable influences $X$ and $Y$, it will introduce terms involving higher-order cumulants of latent variables' noises into the elements of the cumulant matrix. Consequently, it is necessary to extract additional information from the observational data, which correspondingly requires increasing the order of the cumulant matrices. Similar to the Theorem~\ref{th:causal discovery of observed variable}, the identification theorems for the cases involving latent variables are presented as follows.

\begin{theorem}\label{th:causal discovery of one latent variable}
Assume that data over $ X $ and $ Y $ are generated by Eq.~\eqref{eq:path coefficients LvLiNGAM X,Y} and they are influenced by only one latent variable $L$. Define $ CM^{(k)}_{(X,Y)} $ and $ CM^{(k)}_{(Y,X)} $ as cumulant matrices of $ X $ and $ Y $ with $k = 3$. Then, 1) $ X$ is a cause of $Y $ if and only if $rank(CM^{(k)}_{(X,Y)})<rank(CM^{(k)}_{(Y,X)})$; 2) $ X \Vbar Y \mid 
 L$ if and only if $rank(CM^{(k)}_{(X,Y)})=rank(CM^{(k)}_{(Y,X)})=2$.
\end{theorem}

Theorem~\ref{th:causal discovery of one latent variable} guarantees that the rank constraint of the cumulant matrix can identify the causal structure between two observed variables influenced by only one latent variable. 

\begin{theorem}\label{pro:rank constraint of the cumulant matrix}
Assume that $ X $ and $ Y $ are generated by Eq.~\eqref{eq:path coefficients LvLiNGAM X,Y} and they are influenced by $m$ latent variables. Define $ CM^{(k)}_{(X,Y)} $ as cumulant matrix of $ X $ and $ Y $ with $k = m + 2$. Then $rank(CM^{(k)}_{(X,Y)}) = m+p$ if and only if $X$ contains $p$ noises of observed variables.
\end{theorem}

In this paper, we focus on the causal relationship between the two variables, $X$ and $Y$, and thus the value of $p$ can only be 1 or 2. When $p = 1$, it means that $X$ is only influenced by its own noise $E_x$; when $p = 2$, it indicates that $X$ is additionally influenced by the noise $E_y$ of $Y$, implying that $Y$ is a cause for $X$.

\begin{theorem}\label{th:causal discovery of m latent variables}
Assume that $ X $ and $ Y $ are generated by Eq.~\eqref{eq:path coefficients LvLiNGAM X,Y} and they are influenced by $m$ latent variables. Define $ CM^{(k)}_{(X,Y)} $ and $ CM^{(k)}_{(Y,X)} $ as cumulant matrices of $ X $ and $ Y $ with $k = m+2$. Then, 1) $X$ is a cause of $Y$, if and only if $rank(CM^{(k)}_{(X,Y)}) < rank(CM^{(k)}_{(Y,X)})$; 2) $ X \Vbar Y \mid (L_1,\dots,L_m)$ if and only if $rank(CM^{(k)}_{(X,Y)})=rank(CM^{(k)}_{(Y,X)})=m+1$.
\end{theorem}

Theorem~\ref{th:causal discovery of m latent variables} guarantees that the rank constraint of the cumulant matrix can be used to identify causal structures under the influence of an arbitrary number of latent variables.

Theorem~\ref{pro:rank constraint of the cumulant matrix} and Theorem~\ref{th:causal discovery of m latent variables} 
 also indicate that the required order of cumulant matrices varies with the number of latent variables. Notably, by examining the rank of the cumulant matrix, we can determine the number of latent variables that jointly influence two observed variables. The corresponding theorem is presented as follows. 

\begin{theorem}\label{th:rank constraint of latent num}
Assume that $X$ and $Y$ are generated by Eq.~\eqref{eq:path coefficients LvLiNGAM X,Y} and they are influenced by $m$ latent variables. Define $CM^{(k)}_{(X,Y)}$ as the $k$-th order cumulant matrix. Suppose $X$ contains $p$ noise terms of observed variables; if $rank(CM^{(k)}_{(X,Y)}) < k$, then $m = rank(CM^{(k)}_{(X,Y)}) - p$.
\end{theorem}

Theorem~\ref{th:rank constraint of latent num} indicates that when the rank of the cumulant matrix is deficient, its rank can represent the number of latent variables influencing $X$ and $Y$. Specifically, if $X$ is the cause variable, that is, $p = 1$, then the number of latent variables $m = rank(CM^{(k)}_{(X,Y)}) - 1$; if $X$ is influenced by $Y$, that is, $p = 2$, then the number of latent variables $m = rank(CM^{(k)}_{(X,Y)}) - 2$.

\subsection{Learning Algorithm}
In practice, existing methods require computing the eigenvalues or singular values of a matrix to identify its rank~\cite{robin2000tests,kleibergen2006generalized}. However, the higher-order cumulant matrix suffers from large variance in its estimated values, and the distribution spaces of different orders of cumulants are distinct. This makes it difficult to find a stable empirical distribution or parameters to test the rank of the cumulant matrix. Therefore, to effectively identify the causal direction of observed variables influenced by arbitrary latent variables, we utilize the determinant of the cumulant matrix to determine the causal direction, which allows us to directly leverage the information from the entire cumulant matrix. These are guaranteed by the following theorem and corollary.

\begin{theorem}\label{th:determinant-asymmetry-of-m-latent-variables}
Assume that $ X $ and $ Y $ are generated by Eq.~\eqref{eq:path coefficients LvLiNGAM X,Y} and that they are influenced by $m$ latent variables. Define $ CM^{(k)}_{(X,Y)} $ and $ CM^{(k)}_{(Y,X)} $ as cumulant matrices of $ X $ and $ Y $ with $k = m+2$. Then $X$ is a cause of $Y$, if and only if $||CM^{(k)}_{(X,Y)}|| < ||CM^{(k)}_{(Y,X)}||$.
\end{theorem}

\begin{corollary}\label{cor:k > m+2 determinant asymmetry of m latent variables}
Assume that $ X $ and $ Y $ are generated by Eq.~\eqref{eq:path coefficients LvLiNGAM X,Y} and that they are influenced by $m$ latent variables. Define $ CM^{(k)}_{(X,Y)}$ and $CM^{(k)}_{(Y,X)} $ as cumulant matrices of $X$ and $Y$ with $k > m+2$. Then $X$ is a cause of $Y$ if and only if $||CM^{(k)}_{(X,Y)}|| < ||CM^{(k)}_{(Y,X)}||$.
\end{corollary}

Theorem \ref{th:determinant-asymmetry-of-m-latent-variables} establishes the identifiability of causal direction using cumulant matrices of a specific order with $k=m+2$. Corollary~\ref{cor:k > m+2 determinant asymmetry of m latent variables} extends the result in Theorem \ref{th:determinant-asymmetry-of-m-latent-variables}, and shows that when the order of the matrix is greater than $m+2$, the causal direction can still be determined by the determinant of the cumulant matrix. This step has strong implications for the practical usage of the approach. 

Based on the above identifiability theorems, we propose a practical algorithm to identify causal relationships among observed variables from observational data in the presence of an arbitrary number of latent variables. We first identify the number of latent variables $m$ that influence the two observed variables $X$ and $Y$ according to Theorem~\ref{th:rank constraint of latent num}. Specifically, based on the cumulant matrices $CM^{(k)}_{(X,Y)}$ and $CM^{(k)}_{(Y,X)}$ for $X$ and $Y$, we gradually increase the order $k$ of the matrices until both $CM^{(k)}_{(X,Y)}$ and $CM^{(k)}_{(Y,X)}$ are rank-deficient. In such a case, $m = \min(rank(CM^{(k)}_{(X,Y)}), rank(CM^{(k)}_{(Y,X)})) - 1$. Finally, we identify the causal direction between $X$ and $Y$ based on Corollary~\ref{cor:k > m+2 determinant asymmetry of m latent variables}. The specific procedure is summarized in Algorithm~\ref{alg 1}.

\begin{algorithm}[t]
    \caption{Identification of Causal Direction under an Arbitrary Number of Latent Confounders}
    \label{alg 1}
    \begin{multicols}{2}
    \begin{algorithmic}[1]
        \STATE {\bfseries Input:} Data of $X$, $Y$, threshold $\epsilon = 1e-5$.
        \STATE {\bfseries Output:}  Causal direction between $X$ and $Y$.
        \STATE Initialize $k := 2$\;
        \STATE Calculate $CM^{(k)}_{(X,Y)}$, $CM^{(k)}_{(Y,X)}$\;
        \WHILE{$rank(CM^{(k)}_{(X,Y)})$ and $rank(CM^{(k)}_{(Y,X)})=k$}
            \STATE $k := k + 1$\;
            \STATE Update $CM^{(k)}_{(X,Y)}$, $CM^{(k)}_{(Y,X)}$\;
        \ENDWHILE
        \STATE $m := k - 2$\;
        \STATE Calculate $||CM^{(k)}_{(X,Y)}||$ and $||CM^{(k)}_{(Y,X)}||$\; 
        \IF{$abs(||CM^{(k)}_{(X,Y)}|| - ||CM^{(k)}_{(Y,X)}||) < \epsilon$}
            \STATE Infer $ X \Vbar Y \mid (L_1,\dots,L_m)$\;
        \ELSIF{$||CM^{(k)}_{(X,Y)}|| < ||CM^{(k)}_{(Y,X)}||$}
            \STATE Infer $X\rightarrow Y$\;
        \ELSIF{$||CM^{(k)}_{(X,Y)}|| > ||CM^{(k)}_{(Y,X)}||$}
            \STATE Infer $Y\rightarrow X$\;
        \ENDIF 
    \end{algorithmic}
    \end{multicols}
\end{algorithm}

\section{Experiments}

\subsection{Results on Simulated Data}




\paragraph{Experiment Setups.}

In the experiments on simulated data, we considered the following three cases.

\textbf{[Case 1]: } The two observed variables are not affected by latent variables, and there exists a causal directed edge between observed variables, i.e., Fig.~\ref{fig:data generating processes} (a).

\textbf{[Case 2]: } The two observed variables are affected by a single latent variable, and there exists a causal directed edge between observed variables, i.e., Fig.~\ref{fig:data generating processes} (b).

\textbf{[Case 3]: } The two observed variables are affected by more than one latent variable, and there exists a directed edge between observed variables, i.e., Fig.~\ref{fig:data generating processes} (c).

Based on the causal graphs in Fig.~\ref{fig:data generating processes} (a)-(c), we randomly generate data using Eq.~\eqref{eq:LvLiNGAM}, where the causal coefficients are sampled from a uniform distribution within the range $[-0.8,-0.2]\cup[0.2, 0.8]$. The noise terms are generated by applying the logarithm transformation to the absolute value of random numbers generated by the following 5 distributions: \texttt{Laplace}, \texttt{Normal}, \texttt{Logistic}, \texttt{Exponential}, \texttt{Uniform}, denoted as $d_1$, $d_2$, $d_3$, $d_4$ and $d_5$, respectively. This way, they are clearly non-Gaussian. Among these distributions, their scale parameters are sampled from the uniform distribution within the range $[0.5, 1.5]$. 
For each setting, we randomly generate 100 datasets.

We designed experiments for the three cases mentioned above, using five different distributions with sample sizes $N = \{5k, 10k, 50k\}$ to evaluate the performance of different methods in identifying the causal direction. Then, we start with two latent variables in Case 3, and later we will present experimental results involving more than two latent variables. Besides, we also conduct some robustness studies, in which the assumptions are violated, and the results are provided in Appendix.

We compare our method with Overcomplete ICA-based lvLiNGAM (denoted as OL)~\cite{tashiro2012estimation}, LvHC~\cite{chen2024identification}, and ANM~\cite{hoyer2008nonlinear}. To evaluate the performance of different method, we use the accuracy score as an evaluation metric, which is defined as the ratio of correctly identified causal directions to the total number of true causal relationships. All experiments are conducted on a machine equipped with Ubuntu 16.04.5 LTS operating system with an Intel Core i7-6700 CPU, 15GB of RAM.

\paragraph{Evaluation in Case 1.}
As shown in Table~\ref{tab:case 1-3 accuracy score}, our method achieved an accuracy of 1.0 across all distributions, the highest accuracy among all methods. OL achieved an accuracy of 0.99 in distributions $d_1, d_2, d_3, d_4$, but performed worse than LvHC and our method in $d_5$. This is likely due to the fact that OL is more prone to getting stuck in local optima in the specific distribution ($d_5$). The accuracy of LvHC was approximately 0.9, due to its use of redundant higher-order information in Case 1, which increased the identification error. For ANM, its accuracy varied significantly across different distributions. This could be attributed to relying on the HSIC independence test. If the kernel choice of HSIC is not suited, the performance tends to deteriorate. 

\setlength{\tabcolsep}{5pt}
\begin{table}[t]
\begin{center}
\caption{Accuracy scores of ours and baseline methods on Cases 1-3.}
\label{tab:case 1-3 accuracy score}
\begin{tabular}{>{\small}c|>{\small}c|>{\small}c >{\small}c >{\small}c >{\small}c|>{\small}c >{\small}c >{\small}c >{\small}c|>{\small}c >{\small}c >{\small}c >{\small}c}

\hline
\multicolumn{2}{c|}{} & \multicolumn{4}{c|}{Case 1} & \multicolumn{4}{c|}{Case 2} & \multicolumn{4}{c}{Case 3} \\
\hline
$d_i$ & N & Ours & OL & lvHC & ANM & Ours & OL & lvHC & ANM & Ours & OL & lvHC & ANM \\
\hline
     & 5k & \textbf{1.00} & \textbf{1.00} & 0.90 & 0.93 & \textbf{0.93} & 0.87 & 0.92 & 0.08 & \textbf{0.86} & 0.76 & 0.39 & 0.01 \\
$d_1$ & 10k & \textbf{1.00} & \textbf{1.00} & 0.91 & 0.94 & \textbf{0.94} & 0.85 & 0.91 & 0.04 & \textbf{0.85} & 0.77 & 0.34 & 0.00 \\
     & 50k & \textbf{1.00} & 0.99 & 0.96 & - & 0.93 & 0.90 & \textbf{1.00} & - & \textbf{0.83} & 0.79 & 0.44 & - \\
\hline
     & 5k & \textbf{1.00} & 0.96 & 0.92 & 0.92 & \textbf{0.96} & 0.83 & 0.89 & 0.05 & \textbf{0.75} & 0.67 & 0.47 & 0.00 \\
$d_2$ & 10k & \textbf{1.00} & 0.99 & 0.96 & 0.91 & \textbf{0.95} & 0.84 & 0.94 & 0.01 & \textbf{0.80} & 0.72 & 0.37 & 0.00 \\
     & 50k & \textbf{1.00} & 0.99 & 0.98 & - & 0.94 & 0.81 & \textbf{0.96} & - & \textbf{0.81} & 0.78 & 0.38 & - \\
\hline
     & 5k & \textbf{1.00} & 0.97 & 0.87 & 0.96 & \textbf{0.93} & 0.79 & 0.90 & 0.11 & \textbf{0.81} & 0.70 & 0.44 & 0.01 \\
$d_3$ & 10k & \textbf{1.00} & 0.98 & 0.93 & 0.94 & \textbf{0.94} & 0.84 & 0.93 & 0.03 & \textbf{0.80} & 0.79 & 0.45 & 0.00 \\
     & 50k & \textbf{1.00} & 0.98 & 0.97 & - & \textbf{0.96} & 0.82 & 0.95 & - & \textbf{0.79} & 0.77 & 0.41 & - \\
\hline
     & 5k & \textbf{1.00} & 0.94 & 0.86 & 0.98 & \textbf{0.89} & 0.79 & 0.80 & 0.15 & \textbf{0.75} & 0.60 & 0.41 & 0.04 \\
$d_4$ & 10k & \textbf{1.00} & 0.99 & 0.88 & 0.97 & \textbf{0.93} & 0.72 & 0.91 & 0.09 & \textbf{0.72} & 0.68 & 0.42 & 0.01 \\
     & 50k & \textbf{1.00} & 0.97 & 0.92 & - & \textbf{0.94} & 0.75 & \textbf{0.94} & - & \textbf{0.81} & 0.60 & 0.39 & - \\
\hline
     & 5k & \textbf{1.00} & 0.92 & 0.90 & 0.88 & \textbf{0.95} & 0.83 & 0.87 & 0.00 & \textbf{0.76} & 0.71 & 0.41 & 0.00 \\
$d_5$ & 10k & \textbf{1.00} & 0.94 & 0.97 & 0.88 & \textbf{0.95} & 0.81 & 0.93 & 0.00 & \textbf{0.87} & 0.76 & 0.46 & 0.00 \\
     & 50k & \textbf{1.00} & 0.91 & 0.97 & - & 0.94 & 0.82 & \textbf{0.99} & - & \textbf{0.82} & 0.78 & 0.43 & - \\
\hline
\end{tabular}
\end{center}
\end{table}


\paragraph{Evaluation in Case 2.} As shown in Table~\ref{tab:case 1-3 accuracy score}, our method demonstrated the best performance in most cases. For larger sample sizes (50k), the performance of LvHC is slightly better than of ours. This can be attributed to LvHC using higher-order cumulants than ours, which provide more information about the observed variables, but requires a larger sample size for accurate estimation. The performance of OL is inferior to ours and LvHC's. This is because OL requires the assumption of more parameters for identifying latent variables, which increases the dimensionality of its solution space and makes it more prone to local optima. Since Case 2 involves one latent variable, which violates the assumption of ANM, the accuracy of ANM dropped substantially.


\paragraph{Evaluation in Case 3.} As shown in Table~\ref{tab:case 1-3 accuracy score}, our method outperforms all other methods in all settings. As the sample size increases, due to the use of seventh-order cumulants, the estimation of the cumulants becomes more accurate, resulting in improved performance of our method. OL surpasses its identifiable limit, and OICA is more likely to get trapped in local optima under these conditions, resulting in lower accuracy. Both LvHC and ANM perform poorly in Case 3 due to the violation of their assumption.

\begin{wrapfigure}{r}{0.5\textwidth}
    \centering
    \includegraphics[width=0.5\textwidth]{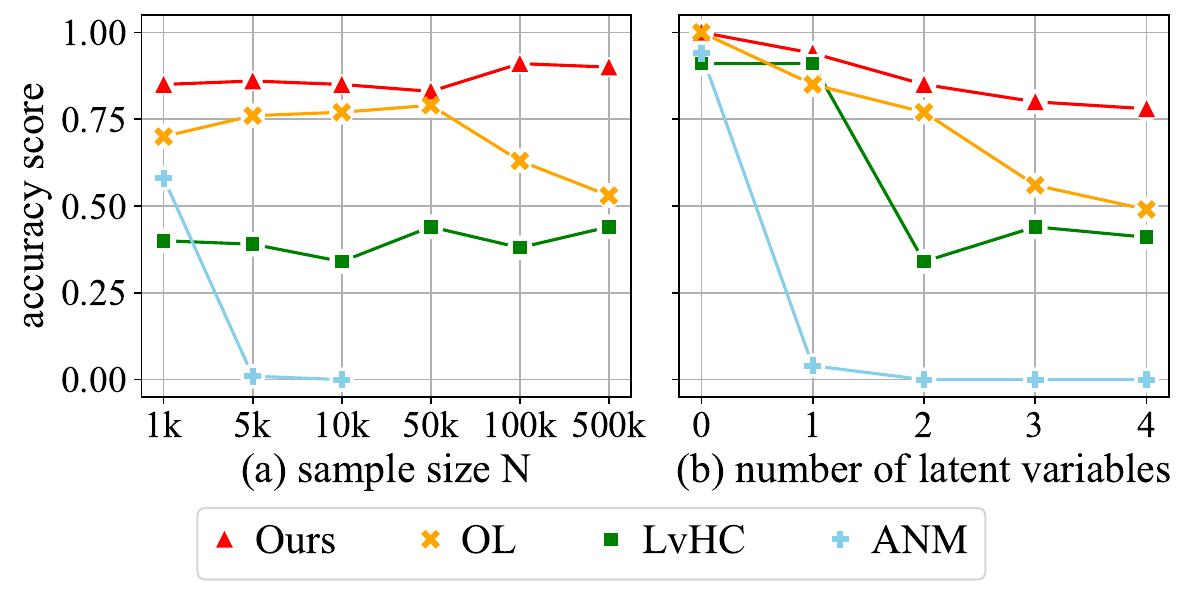}
    \caption{Accuracy scores on various sample sizes and different numbers of latent variables.}
    \label{fig:Various Sample Sizes and Latent Num case}
\end{wrapfigure}

\paragraph{Sensitivity to the sample size and the number of latent variables.} Furthermore, for Case 3, we conducted additional experiments with sample sizes varied in $ \{1k, 5k, 10k, 50k, 100k, 500k\}$ and the number of latent variables varied among $\{0,1,2,3,4\}$. As shown in Fig.~\ref{fig:Various Sample Sizes and Latent Num case} (a), our method 
achieves an accuracy above 0.8. As the sample size increases, the accuracy of our method improves. For OL, increasing sample size doesn't improve accuracy as the number of latent variables exceeds its identifiable limit. Similarly, the setting in Case 3 violates the identifiability assumption of LvHC and ANM, so their accuracy does not improve even under a large sample size. Fig.~\ref{fig:Various Sample Sizes and Latent Num case} (b) illustrates the results under different numbers of latent variables. From the results, our method consistently achieves good performance. In contrast, OL experiences a significant decrease when the number of latent variables exceeds 2. LvHC and ANM only perform well when their assumptions are met.

\begin{wrapfigure}{r}{0.52\textwidth}
  \centering
  \includegraphics[width=0.52\textwidth]{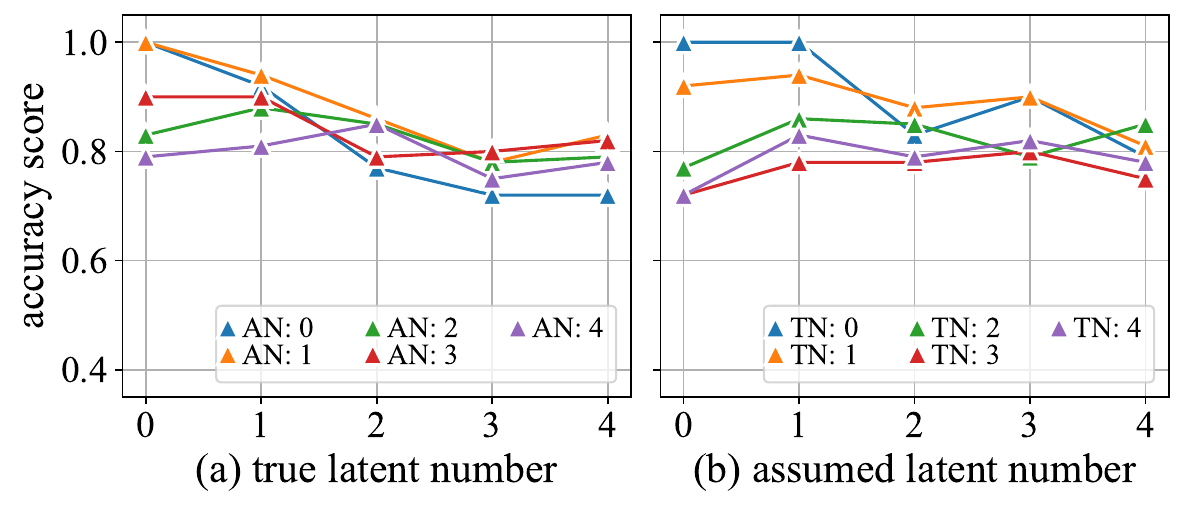}
  \caption{Accuracy scores on various True Number (TN) and Assumed Number (AN) of latent variables.}
  \label{fig:True Number and Assumed Number of Latent Variables case}
\end{wrapfigure}

\paragraph{Sensitivity to the assumed number of latent variables.}
We conducted a series of experiments to evaluate the accuracy of our method in identifying the causal direction of observed variables under different assumed latent variables. Specifically, the true number of latent variables (TN) is varied in $\{0, 1, 2, 3, 4\}$, while the assumed number of latent variables (AN) is varied in $\{0, 1, 2, 3, 4\}$. As shown in Fig.~\ref{fig:True Number and Assumed Number of Latent Variables case} (a), for different true numbers of latent variables, specifically when the true number of latent variables is 0, 1, 2, and 3, the performance in identifying causal directions is better when the assumed number of latent variables equals the true number, compared to most other assumed latent variable scenarios. However, when the number of latent variables is 4, the performance deteriorates, likely due to the larger estimation errors of higher-order cumulants under the same sample size. Therefore, when the true number of latent variables is accurately identified, our method performs the best, with accuracy rates almost always exceeding 0.8. 

Fig.~\ref{fig:True Number and Assumed Number of Latent Variables case} (b) illustrates the performance of our method by setting different assumed numbers of latent variables under different true numbers of latent variables. As shown in Fig.~\ref{fig:True Number and Assumed Number of Latent Variables case} (b), the accuracy of causal direction identification is higher when the number of assumed latent variables equals the true number than that in other cases. When the assumed number of latent variables is greater than the true number, the performance is worse, likely due to the larger estimation variance of higher-order cumulants, leading to increased identification errors. When the assumed number of latent variables is smaller than the true number, the performance also deteriorates, as the number of latent variables exceeds the upper limit of what the cumulant matrix can identify. However, overall, the accuracy for all three scenarios is above 0.8, and only in the extreme case where the true number of latent variables is large (3 or 4) and the assumed number is equal to zero, does the accuracy decline to approximately 0.7. These results imply that our method consistently achieves good accuracy across different assumptions regarding the number of latent variables.

\subsection{Results on Real-World Stock Market Data}

We applied our method and baseline methods to see the causal direction across all variable pairs among the selected 14 stocks in the Hong Kong Stock Market Data~\cite{zhang2008minimal}. We used 6252 daily returns from Jan. 25, 2000, to Jan. 25, 2025, obtained from the Yahoo finance database. More details about the data and the results are provided in Appendix.

From the results, we found that the identified causal relationship of company stock returns by our method is typically directly related to the company's position in the supply chain~\cite{mitra2008supply,lanier2019supply,li2021information}. In detail, 0002.HK, 0003.HK, and 0006.HK rank in the top three, all of which are energy or material companies. These corresponding companies are always the first to receive production orders in production activities.


Surprisingly, we further found that causal relationships also exist between the returns of some stock returns with ownership relationships, which is consistent with previous studies~\cite{zhang2008minimal,imam2007firm,rasyid2015effects,thomsen2006blockholder}. In these data, there are four stock returns with ownership relationships, which are 0001.HK ($X_{1}$), 0010.HK ($X_{5}$), 0011.HK ($X_{8}$), and 0010.HK ($X_{10}$). The ownership relationships among them are: $X_{5}$ holds 60\% of $X_{8}$'s shares, and $X_{1}$ holds 50\% of $X_{10}$'s shares. The causal relationships identified by our algorithm and OL method are: $X_{8} \rightarrow X_{5}$ and $X_{10} \rightarrow X_{1}$. LvHC fails to identify the causal direction between $X_{10}$ and $X_{1}$, and ANM fails to identify both.


The above findings suggest that our method avoids the interference of latent variables in the financial system and can accurately identify the causal directions between stock returns.

\section{Conclusion}

In this paper, leveraging higher-order cumulants, we construct the cumulant matrix and provide the identifiability theory for inferring the causal relationship between two observed variables that are influenced by an arbitrary number of latent variables. Based on the identifiability theory, we developed a practical causal discovery method to identify the number of latent variables and determine the causal direction between the observed variables using the determinant of the cumulant matrix.
The experimental results showed that our proposed method performs almost perfectly with larger sample sizes (around 100,000), supporting the asymptotic correctness of the result. Notably, our method still performs well with smaller sample sizes (ranging from 1,000 to 50,000), indicating that the algorithm we designed can effectively mitigate the impact of errors when estimating the joint cumulants. In summary, the proposed method makes a significant advancement in the field of causal discovery, even in the presence of latent variables. It enhances the transparency and interpretability of complex data-generating mechanisms, which aids in the reduction of bias and is beneficial for decision-making. However, our proposed method requires some mild assumptions that might not be met in real-world scenarios. Therefore, how to relax the assumptions, i.e., linear, non-Gaussian assumption, would be the focus of our future research.

\bibliographystyle{unsrt}
\bibliography{neurips_2025}


\end{document}